\documentclass{svjour3}                     
\smartqed  
\usepackage{graphicx,natbib,multirow,amsmath}
\usepackage{setspace}
\usepackage{url}
\usepackage{multicol}
\usepackage{algorithmic,algorithm}

\begin{document}

\title{HyFlex: A Benchmark Framework for Cross-domain Heuristic Search
}

\titlerunning{HyFlex}        

\author{Edmund Burke \and Tim Curtois \and Matthew Hyde \and Gabriela Ochoa \and Jos\'{e} A. V\'azquez-Rodr\'iguez }

\authorrunning{Burke et al.t} 

\institute{Automated Scheduling, Optimisation and Planning (ASAP) research group  \at
              University of Nottingham, School of Computer Science, \\
              Jubilee Campus, Wollaton Road, Nottingham NG8 1BB, UK\\
              \email{\{ekb,tec,mvh,gxo\}@cs.nott.ac.uk}           
}

\date{Received: date / Accepted: date}

\maketitle


\begin{abstract}

Automating the design of heuristic search methods is an active research field within computer science, artificial intelligence and operational research. In order to make these methods more generally applicable, it is important to eliminate or reduce the role of the human expert in the process of designing an effective methodology to solve a given computational search problem. Researchers developing  such methodologies are often constrained on the number of problem domains on which to test their adaptive, self-configuring algorithms; which can be explained by the inherent difficulty of implementing their corresponding domain specific software components.

This paper presents  {\em HyFlex}, a software framework for the development of cross-domain search methodologies. The framework features a  common software interface for dealing with different combinatorial optimisation problems, and  provides the algorithm components that are problem specific. In this way, the algorithm designer does not require a detailed  knowledge the problem domains, and thus can concentrate his/her efforts  in designing adaptive general-purpose heuristic search algorithms. Four hard combinatorial problems are fully implemented (maximum satisfiability, one dimensional bin packing, permutation flow shop and personnel scheduling), each containing a varied set of instance data (including real-world industrial applications) and an extensive set of problem specific heuristics and search operators. The framework forms the basis for the first International Cross-domain Heuristic Search Challenge (CHeSC), and it is currently in use by the international research community. In summary, HyFlex represents a valuable new benchmark of heuristic search generality, with which adaptive cross-domain algorithms are being easily developed, and reliably compared.

\end{abstract}

\section{Introduction}
\label{intro}

There is a renewed and growing research interest in techniques for automating the design of heuristic search methods. The goal is to remove or reduce the need for a human expert in the process of designing an effective algorithm to solve a search problem, and consequently raise the level of generality at which search methodologies can operate.  Evolutionary algorithms and  metaheuristics have been successfully applied to solve a variety of real-world complex optimisation problems. Their design, however, has become increasingly complex.  In order to make these methodologies widely applicable, it is important to provide self-managed systems that can configure themselves `on the fly'; adapting to the changing problem (or search space) conditions, based on general high-level guidelines provided by their users.

Researchers pursuing these goals within combinatorial optimisation, are often limited by the number of problems domains available to them for testing their adaptive methodologies. This can be explained by the difficulty and effort required to implement state-of-the-art software components, such as the problem model, solution representation, objective function evaluation and search operators;  for many different combinatorial optimisation problems. Although several benchmark problems in combinatorial optimisation are available \citep{Taillard93,SATcomp,esicup,ORLIB,TSPLIB} (to name just a few); they contain mainly the data of a set of instances and their best known solutions. They generally do not incorporate the software necessary to encode the solutions and calculate the objective function, let alone existing search operators for the given problem. It is the researcher who needs to provide these in order to later test their high-level adaptive search method.  To overcome such limitations, we propose {\em HyFlex}, a modular and flexible Java class library for designing and testing iterative heuristic search algorithms.  It provides a number of problem domain modules, each of which encapsulates the problem-specific algorithm components: solution representation, fitness evaluation, instance data, and a repository of associated problem-specific heuristics. Importantly, only the high-level control strategy needs to be implemented by the user, as HyFlex provides an easy to use interface with which the problem domains can be accessed. Indeed, HyFlex can be considered as an extension of the notion of a benchmark for combinatorial optimisation. Instead of providing only a data-set for a given problem domain, HyFlex also provides the problem specific software surrounding it. Thus, HyFlex acts as a benchmark for cross-domain optimisation and more general search methodologies.

A number of techniques and research themes within operational research, computer science and artificial intelligence would benefit from the proposed framework. Among them: hyper-heuristics \citep{burke03tabuhh,hhchap03,Burke2010,Ross:05}, adaptive memetic algorithms \citep{KraSmi:00,Jakob:06,Ong2006,Smith2007,Neri2007},  adaptive operator selection  \citep{Fialho2008,Fialho2010,Maturana2008,Maturana2010}, reactive search \citep{Battiti:96,Battiti2009}, variable neighborhood search \citep{Mladenovic:97} and its adaptive variants \citep{Braysy2002,Ropke07}; and generally the development of adaptive parameter control strategies in evolutionary algorithms  \citep{Eiben2007,Lobo2007}.  HyFlex can be seen, then, as a unifying benchmark, with which the performance of different adaptive techniques can be reliably assessed and compared. Indeed, HyFlex is currently used to support an international research competition: the First Cross-Domain Heuristic Search Challenge \citep{chesc2011}. The challenge is analogous to the athletics Decathlon event, where the goal is not to excel in one event at the expense of others, but to have a good general performance on each. The competition will also provide a set of state-of-the-art initial results on the HyFlex benchmark. Competitors will submit one Java class file representing their hyper-heuristic or high-level search strategy. This class file will then be run in HyFlex through the common interface. This ensures that the competition is fair, because all of the competitors must use the same problem representation and search operators. Moreover, due to the common interface, the competition will consider not only hidden instances, but also hidden domains. An interesting feature of CHeSC is the Leaderboard, a table which ranks participants according to their best score on a rehearsal competition conducted every week. This rehearsal competition is based on a set of results submitted by the participants who chose to do so. It has brought substantial dynamism and interest to the challenge.  CHeSC currently has 43 registered teams from 23 different countries.

This article is structured as follows. Section \ref{hyflex} describes the antecedents and architecture of the HyFlex framework. It also includes examples of how to implement and run hyper-heuristics within the framework. Section \ref{domains} presents the four problem domains which are currently implemented:  maximum satisfiability (MAX-SAT), one-dimensional bin packing, permutation flow shop, and personnel scheduling. For each domain, details are given on the instance data, solution initialisation method, objective function evaluation, and the set of problem specific heuristics. Section \ref{algorithms} illustrates the implementation of three high-level search strategies using HyFlex: an iterative hyper-heuristic, a multiple neighbourhood iterated local search algorithm, and a multi-meme memetic algorithm.  They are not intended to be state-of-the-art adaptive approaches in their categories. Instead, they were selected to illustrate the wide range of algorithm designs that can be implemented within HyFlex. Section \ref{experiments} presents a comparative study of the three algorithms. The goal is not to determine the best performing algorithm, but instead to illustrate their difference in behavior across the different problem domains. Finally, section \ref{conclusions} summarises our contribution and suggests directions for future research.

\section{The HyFlex Framework}
\label{hyflex}

\subsection{Overview of HyFlex}\label{overview}
\begin{figure}
  \centering
  \includegraphics[scale=1]{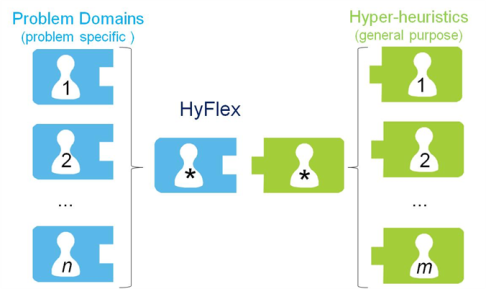}
  \caption{Modularity of heuristic search algorithms. Separation between the problem-specific and the general-purpose parts, both of which are reusable and interchangeable through the HyFlex interface.}
  \label{HyFlex_modular}
\end{figure}

HyFlex (Hyper-heuristics Flexible framework) is a software framework designed to enable the development, testing and comparison of iterative general-purpose heuristic search algorithms (such as hyper-heuristics). To achieve these goals it uses modularity and the concept of decomposing a heuristic search algorithm into two main parts (see Figure \ref{HyFlex_modular}):

\begin{enumerate}
\item A general-purpose part: the algorithm or hyper-heuristic.
\item The problem-specific part: provided by the HyFlex framework.
\end{enumerate}

\noindent
In the hyper-heuristics literature, this idea is also referred to as the domain barrier between the problem-specific heuristics and the hyper-heuristic \citep{hhchap03,cowling00}. HyFlex extends the conceptual domain-barrier framework  by maintaining a population  (instead of a single incumbent solution)  in the problem domain layer. Moreover, a richer variety of problem specific heuristics and search operators is provided. Another relevant antecedent to HyFlex is PISA \citep{pisa03}, a text-based software interface for multi-objective evolutionary algorithms. PISA provides a division between the application-specific and the algorithm-specific parts of a multi-objective evolutionary algorithm. In HyFlex, the interface is not text-based. Instead, it is  given by an abstract Java class. This allows a more tight coupling between the modules and overcomes some of the speed limitations encountered in  PISA. While PISA is designed to implement evolutionary algorithms, HyFlex can be used to implement  both population-based and single point metaheuristics and hyper-heuristics.  Moreover, it provides a rich variety of fully implemented combinatorial optimisation problems including real-world instance data.

\noindent
The framework is written in java which is familiar to and commonly used by many researchers. It also benefits from object orientation, platform independence and automatic memory management. At the highest level the framework consists of just two abstract classes: ProblemDomain and HyperHeuristic. The structure of these classes is shown in the class diagram of figure \ref{UML_diagram}. In the diagram, the signatures adjacent to circles are public methods and fields, and the signatures adjacent to diamonds are protected. Abstract methods are denoted by italics, and the implementations of these methods are necessarily different for each problem domain class.

\subsubsection{The \texttt{ProblemDomain }Class}\label{sec:probdomain}
\begin{figure}
  \centering
  \includegraphics[scale=0.6]{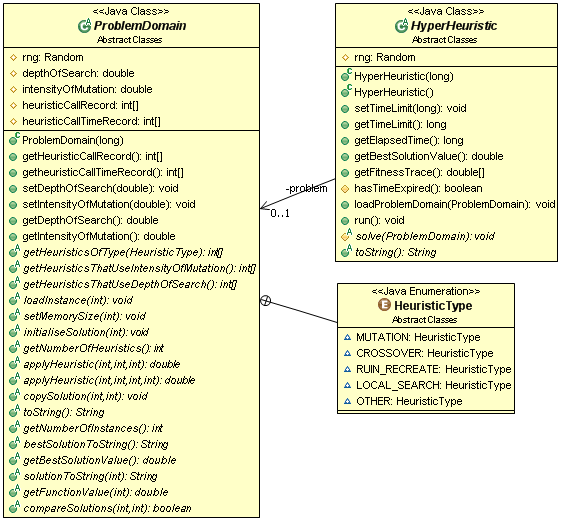}
  \caption{Class diagram for the HyFlex framework.}
  \label{UML_diagram}
\end{figure}

As shown in figure \ref{UML_diagram}, an implementation of the \texttt{ProblemDomain} class provides the following elements, each of which is easily accessed and managed with one or more methods.

\begin{enumerate}
\item A user-configurable memory (a population) of solutions, which can be managed by the hyper-heuristic through methods such as \texttt{setMemorySize} and \texttt{copySolution}.
\item A routine to randomly initialise solutions, \texttt{initialiseSolution($i$)}, where $i$ is the index of the solution index in the memory.\subsubsection{Description}
\noindent
{\bf Problem formulation}: `SAT' refers to the boolean satisfiability problem. This problem involves determining if there is an assignment of the boolean variables of a formula, which results in the whole formula evaluating to true. If there is such an assignment then the formula is said to be satisfiable, and if not then it is unsatisfiable.
An example formula is given in equation \ref{eqn1}, which is satisfied when $x_1 = false$ $x_2 = false$ $x_3 = true$ and $x_4 = false$.
\begin{equation}
\label{eqn1}
(x_1 \vee \neg x_2 \vee \neg x_3) \wedge (\neg x_1 \vee x_3 \vee x_4) \wedge (x_2 \vee \neg x_3 \vee \neg x_4)
\end{equation}

HyFlex implements  one of SAT's related optimisation problems,  the maximum satisfiability problem (MAX-SAT), in which the objective is to find the maximum number of clauses of a given Boolean formula that can be satisfied by some assignment. The problem can also be formulated as a minimisation problem, where the objective is to minimise the number of unsatisfied clauses.

\noindent
{\bf Solution initialisation}: The solutions are initialised by  randomly assigning a true or false value to each variable.

\noindent
{\bf Objective function}: The fitness function returns the number of `broken' clauses, which are those which evaluate to false.

\noindent
{\bf Instance data}: The ten training instances and their sources are summarised in Table \ref{tab:satinst}.
\begin{table}
\centering
\caption{MAX-SAT instances}
\begin{tabular}{c|c|c|c|c}
& name & source & variables & clauses\\
\hline
1 & {\tiny contest02-Mat26.sat05-457.reshuffled-07} & {\small \cite{SAT07}} & 744 & 2464 \\
2 & {\tiny hidden-k3-s0-r5-n700-01-S2069048075.sat05-488.reshuffled-07} & {\small \cite{SAT07}} & 700 & 3500   \\
3 & {\tiny hidden-k3-s0-r5-n700-02-S350203913.sat05-486.reshuffled-07}  & {\small \cite{SAT07}}&  700&  3500 \\
4 & {\tiny parity-games/instance-n3-i3-pp}   & {\small \cite{SAT09}} &  525 & 2276 \\
5 & {\tiny parity-games/instance-n3-i3-pp-ci-ce} & {\small \cite{SAT09}}&  525 & 2336 \\
6 & {\tiny parity-games/instance-n3-i4-pp-ci-ce} & {\small \cite{SAT09}} & 696 & 3122 \\
7 & {\tiny highgirth/3SAT/HG-3SAT-V250-C1000-1} & {\small \cite{SATcomp}} & 250 & 1000\\
8 & {\tiny highgirth/3SAT/HG-3SAT-V250-C1000-2} & {\small \cite{SATcomp}} &  250 & 1000\\
9 & {\tiny highgirth/3SAT/HG-3SAT-V300-C1200-2} & {\small \cite{SATcomp}} &  300 & 1200\\
10 & {\tiny MAXCUT/SPINGLASS/t7pm3-9999}        & {\small \cite{SATcomp}} & 343 & 2058 \\
\end{tabular}
\label{tab:satinst}
\end{table}

\subsubsection{Search Operators}
This domain contains a total of 9 search operators, summarised by \citet{fukunaga08}. Before describing them, find below four relevant definitions. Let $T$ be the state of the formula before the variable is flipped, and let $T'$ be the state of the formula after the variable is flipped.
\begin{description}
\item {\bf Net gain} of a variable is defined as the number of broken clauses in T minus the number of broken clauses in $T'$.
\item {\bf Positive gain} of a variable is the number of broken clauses in $T$ that are satisfied in $T'$.
\item {\bf Negative gain} of a variable is the number of satisfied clauses in $T$ that are broken in $T'$.
\item {\bf Age} of a variable is the number of variable flips since it was last flipped.
\end{description}
\begin{description}
\item{\bf Mutational heuristics}
\begin{itemize}
    \item[$h_1$:] GSAT: Flip the variable with the highest net gain, and break ties randomly \citep{selman92}.
    \item[$h_2$:] HSAT: Identical functionality to GSAT, but ties are broken by selecting the variable with the highest age \citep{gent93}.
    \item[$h_3$:] WalkSAT: Select a random broken clause BC. If any variables in BC have a negative gain of zero, randomly select one of these to flip. If no such variable exists, flip a random variable in BC with probability 0.5, otherwise flip the variable with minimal negative gain \citep{selman94}.
    \item[$h_4$:] Novelty: Select a random broken clause BC. Flip the variable v with the highest net gain, unless v has the minimal age in BC. If this is the case, then flip it with 0.3 probability. Otherwise flip the variable with the second highest net gain \citep{mcallester97}.
\end{itemize}
\item{\bf Ruin-recreate heuristics}
\begin{itemize}
    \item[$h_5$:] A proportion of the variables is randomly reinitialised.
\end{itemize}

\item{\bf Local search heuristics}
\begin{itemize}
    \item[$h_6$:] This is a first-improvement local search. In each iteration, flip a variable selected completely at random.
    \item[$h_7$:] This is a first-improvement local search. In each iteration, flip a randomly selected variable from a randomly selected broken clause.
\end{itemize}

\item{\bf Crossover heuristics}
\begin{itemize}
    \item[$h_8$:] Standard one point crossover on the boolean strings of variables.
    \item[$h_9$:] Standard two point crossover on the boolean strings of variables.
\end{itemize}
\end{description}

\item A set of problem specific heuristics, which are used to modify solutions. These are called by the user's hyper-heuristic with the \texttt{applyHeuristic($i,j,k$)} method, where $i$ is the index of the heuristic to call, $j$ is the index of the solution in memory to modify, and $k$ is the index in memory where the resulting solution should be put. Each problem-specific heuristic in each problem domain is classified into one of four groups, shown below. The heuristics belonging to a specific group can be accessed by calling \texttt{getHeuristicsOfType($type$)}.
  \begin{itemize}
  \item Mutational or perturbation heuristics: perform a small change on the solution, by swapping, changing, removing, adding or deleting solution components.
  \item Ruin-recreate (destruction-construction) heuristics: partly destroy the solution and rebuild or recreate it afterwards. These heuristics can be considered as large neighbourhood structures. They are, however, different from the mutational heuristics in that they can incorporate problem specific construction heuristics to rebuild the solutions
  \item Hill-climbing or local search heuristics: iteratively make small changes to the solution, only accepting non-deteriorating solutions, until a local optimum is found or a stopping condition is met. These heuristics differ from mutational heuristics in that they incorporate an iterative improvement process, and they guarantee that a non-deteriorating solution will be produced.
  \item Crossover heuristics: take two solutions, combine them, and return a new solution.
\end{itemize}
\item A varied set of instances that can be easily loaded using the method \texttt{loadInstance($a$)}, where $a$ is the index of the instance to be loaded.
\item A fitness function, which can be called with the \texttt{getFunctionValue($i$)} method, where $i$ is the index of the required solution in the memory. HyFlex problem domains are always implemented as minimisation problems, so a lower fitness is always superior. The fitness of the best solution found so far in the run can be obtained with the \texttt{getBestSolutionValue()} method.
\item Two parameters: $\alpha$ and $\beta, (0 <= [\alpha, \beta] <= 1)$, which are the `intensity' of mutation and `depth of search', respectively, that control the behaviour of some search operators.
\end{enumerate}

\subsubsection{The \texttt{HyperHeuristic} Class}
The HyperHeuristic class is designed to allow algorithms which implement this class to be compared and benchmarked across one or more of the problem domains available (for example, in a competition). Users create cross-domain heuristic algorithms by creating implementations of this abstract class. Each class must contain a \texttt{toString()} method, to give the methodology a name. It must also contain a \texttt{solve()} method, in which the functionality of the particular methodology is written.

The \texttt{solve()} method would normally contain a loop, which continues while the time limit (defined by the user) has not been exceeded. In the loop, the code should provide a mechanism for selecting between the available problem-specific heuristics, and choose to which solutions in memory to apply the heuristics. This class could choose to work with a memory size of 1 for a single point search, or a large memory could be maintained for a population based approach. The memory can be easily defined and maintained through calling methods of the \texttt{ProblemDomain} class, where the memory is stored. A hyper-heuristic class automatically records the length of time for which it has been running, and this can be monitored through methods such as \texttt{hasTimeExpired()} and \texttt{getElapsedTime()}.

The solve method is the only method which must be implemented, all other common functionality is provided by the HyFlex software, such as the timing function and the recording of the best solution.

\subsection{Running a Hyper-Heuristic}\label{rhh}
Algorithm \ref{alg:run} shows the ease with which a hyper-heuristic can be run on a problem domain.
An object is created for the problem domain (in this example MAX-SAT), and for the hyper-heuristic, each with a random seed. Then a problem instance is loaded from the selection available in the problem domain object. In this example we choose the instance with index 0. The problem domain is now set up for the hyper-heuristic.

We set the time for which the hyper-heuristic will run, in milliseconds. Then the hyper-heuristic object is given a reference to the problem domain object. Now that the setup is complete, the \texttt{run()} method of the hyper-heuristic is called, to start the search process. The hyper-heuristic will run for 60 seconds in this example, and the best solution found during that time is retrievable with the \texttt{getBestSolutionValue()} method, as shown in algorithm \ref{alg:run}.

\begin{algorithm}
\caption{{Java code for running a hyper-heuristic on a problem domain}}
\label{alg:run}
\begin{algorithmic}
\STATE ProblemDomain problem = new SAT(1234);
\STATE HyperHeuristic HHObject = new ExampleHyperHeuristic1(5678);
\STATE problem.loadInstance(0);
\STATE HHObject.setTimeLimit(60000);
\STATE HHObject.loadProblemDomain(problem);
\STATE HHObject.run();
\STATE System.out.println(HHObject.getBestSolutionValue());
\end{algorithmic}
\end{algorithm}

\subsection{An Example Hyper-Heuristic}
This section provides an example hyper-heuristic, to illustrate the ease with which a hyper-heuristic can be created. This is done by extending the HyperHeuristic abstract class, and implementing only one method. All of the common functionality is provided by the HyFlex software, such as the timing function and the recording of the best solution. This example demonstrates exactly how to use certain elements of HyFlex functionality, including the solution memory.

After the \texttt{run()} method of the hyper-heuristic is called (see section \ref{rhh}), the hyper-heuristic abstract class performs some housekeeping tasks, such as initialising the timer, and then calls the \texttt{solve} method of the chosen hyper-heuristic. In our example this is an object of the class \texttt{ExampleHyperHeuristic1}. Algorithm \ref{alg:hh} shows the code for the \texttt{solve()} method in \texttt{ExampleHyperHeuristic1}. It shows that very few lines of code are necessary in order to implement a hyper-heuristic method with the HyFlex framework. Algorithm \ref{alg:hh} is written in pseudocode, but each line corresponds to no more than one line of actual Java code. The \texttt{solve()} method is the only substantial method which needs to be implemented. Indeed the only other necessary method is \texttt{toString()}, which requires one line to give the hyper-heuristic a name.

From Algorithm \ref{alg:hh}, we can see that the \texttt{solve()} method takes the problem domain object as an argument, and first checks for the number of search operators available within it. We also initialise a value to store the current objective function value. It is also necessary to initialise at least one solution in the memory. The default memory size is 2, and we initialise the solution at index 0, which means we build an initial solution with the method specified in the problem domain (generally a fast randomised constructive heuristic). The solution at index 1 remains uninitialised, and therefore has a value of \texttt{null}.

An implemented hyper-heuristic must always contain a while loop which checks if the time limit has expired. The code within the loop specifies the main functionality of the hyper-heuristic. In this example, we choose a random operators, and then apply it to the solution at index 0. The modified solution is put in the memory at index 1 (previously not initialised). Note that a random number generator \texttt{rng} is provided by the HyperHeuristic abstract class. This is created when the hyper-heuristic object's constructor is called, and is the reason why that constructor requires a random seed.

If the new solution is superior to the old solution, it is accepted, and the new solution overwrites the old one in memory. The \texttt{copySolution} method of the problem domain class is employed to manage this. If the new solution is not superior, then the new solution is accepted with 0.5 probability.

\begin{algorithm}
\caption{{Pseudocode for the solve method of \texttt{ExampleHyperHeuristic1}. This is called when the \texttt{run()} method of the hyper-heuristic is called (see algorithm \ref{alg:run})}}
\label{alg:hh}
\begin{algorithmic}
\REQUIRE A ProblemDomain object, problem
\STATE 		int numberOfHeuristics = problem.getNumberOfHeuristics
\STATE 		double currentObjValue = Double.POSITIVE-INFINITY
\STATE 		problem.initialiseSolution(0)
\WHILE		{hasTimeExpired = FALSE}
\STATE 			int h = rng.nextInt(numberOfHeuristics)
\STATE 			double newObjValue = problem.applyHeuristic(h, 0, 1)
\STATE 			double $delta$ = currentObjValue - newObjValue
\IF 			{$delta$ $>$ 0}
\STATE 				problem.copySolution(1, 0)
\STATE 				currentObjValue = newObjValue;
\ELSE 			
\IF 				{rng.nextBoolean = TRUE}
\STATE 					problem.copySolution(1, 0)
\STATE 					currentObjValue = newObjValue;
\ENDIF
\ENDIF
\ENDWHILE
\end{algorithmic}
\end{algorithm}

\subsection{Summary of HyFlex Description}
In this section, we have given an overview of the HyFlex framework, and demonstrated that it is very easy to create and run a hyper-heuristic using the framework. The contribution of HyFlex is that the hyper-heuristic developer now does not need expertise in any of the problem domains. The developer is therefore free to focus their research efforts into developing hyper-heuristic methodologies which can be shown to be generally successful across a range of problem domains.

\section{HyFlex Problem Domains}
\label{domains}
Currently, four problem domain modules are implemented(which can be downloaded from \cite{chesc2011}):  maximum satisfiability (MAX-SAT), one-dimensional bin packing, permutation flow shop, and personnel scheduling. Each domain includes 10 training instances from different sources, and number of problem-specific heuristics of the types discussed in section \ref{overview}.

\subsection{Maximum Satisfiability (MAX-SAT)}
\label{maxsat}

\subsubsection{Description}
\noindent
{\bf Problem formulation}: `SAT' refers to the boolean satisfiability problem. This problem involves determining if there is an assignment of the boolean variables of a formula, which results in the whole formula evaluating to true. If there is such an assignment then the formula is said to be satisfiable, and if not then it is unsatisfiable.
An example formula is given in equation \ref{eqn1}, which is satisfied when $x_1 = false$ $x_2 = false$ $x_3 = true$ and $x_4 = false$.
\begin{equation}
\label{eqn1}
(x_1 \vee \neg x_2 \vee \neg x_3) \wedge (\neg x_1 \vee x_3 \vee x_4) \wedge (x_2 \vee \neg x_3 \vee \neg x_4)
\end{equation}

HyFlex implements  one of SAT's related optimisation problems,  the maximum satisfiability problem (MAX-SAT), in which the objective is to find the maximum number of clauses of a given Boolean formula that can be satisfied by some assignment. The problem can also be formulated as a minimisation problem, where the objective is to minimise the number of unsatisfied clauses.

\noindent
{\bf Solution initialisation}: The solutions are initialised by  randomly assigning a true or false value to each variable.

\noindent
{\bf Objective function}: The fitness function returns the number of `broken' clauses, which are those which evaluate to false.

\noindent
{\bf Instance data}: The ten training instances and their sources are summarised in Table \ref{tab:satinst}.
\begin{table}
\centering
\caption{MAX-SAT instances}
\begin{tabular}{c|c|c|c|c}
& name & source & variables & clauses\\
\hline
1 & {\tiny contest02-Mat26.sat05-457.reshuffled-07} & {\small \cite{SAT07}} & 744 & 2464 \\
2 & {\tiny hidden-k3-s0-r5-n700-01-S2069048075.sat05-488.reshuffled-07} & {\small \cite{SAT07}} & 700 & 3500   \\
3 & {\tiny hidden-k3-s0-r5-n700-02-S350203913.sat05-486.reshuffled-07}  & {\small \cite{SAT07}}&  700&  3500 \\
4 & {\tiny parity-games/instance-n3-i3-pp}   & {\small \cite{SAT09}} &  525 & 2276 \\
5 & {\tiny parity-games/instance-n3-i3-pp-ci-ce} & {\small \cite{SAT09}}&  525 & 2336 \\
6 & {\tiny parity-games/instance-n3-i4-pp-ci-ce} & {\small \cite{SAT09}} & 696 & 3122 \\
7 & {\tiny highgirth/3SAT/HG-3SAT-V250-C1000-1} & {\small \cite{SATcomp}} & 250 & 1000\\
8 & {\tiny highgirth/3SAT/HG-3SAT-V250-C1000-2} & {\small \cite{SATcomp}} &  250 & 1000\\
9 & {\tiny highgirth/3SAT/HG-3SAT-V300-C1200-2} & {\small \cite{SATcomp}} &  300 & 1200\\
10 & {\tiny MAXCUT/SPINGLASS/t7pm3-9999}        & {\small \cite{SATcomp}} & 343 & 2058 \\
\end{tabular}
\label{tab:satinst}
\end{table}

\subsubsection{Search Operators}
This domain contains a total of 9 search operators, summarised by \citet{fukunaga08}. Before describing them, find below four relevant definitions. Let $T$ be the state of the formula before the variable is flipped, and let $T'$ be the state of the formula after the variable is flipped.
\begin{description}
\item {\bf Net gain} of a variable is defined as the number of broken clauses in T minus the number of broken clauses in $T'$.
\item {\bf Positive gain} of a variable is the number of broken clauses in $T$ that are satisfied in $T'$.
\item {\bf Negative gain} of a variable is the number of satisfied clauses in $T$ that are broken in $T'$.
\item {\bf Age} of a variable is the number of variable flips since it was last flipped.
\end{description}
\begin{description}
\item{\bf Mutational heuristics}
\begin{itemize}
    \item[$h_1$:] GSAT: Flip the variable with the highest net gain, and break ties randomly \citep{selman92}.
    \item[$h_2$:] HSAT: Identical functionality to GSAT, but ties are broken by selecting the variable with the highest age \citep{gent93}.
    \item[$h_3$:] WalkSAT: Select a random broken clause BC. If any variables in BC have a negative gain of zero, randomly select one of these to flip. If no such variable exists, flip a random variable in BC with probability 0.5, otherwise flip the variable with minimal negative gain \citep{selman94}.
    \item[$h_4$:] Novelty: Select a random broken clause BC. Flip the variable v with the highest net gain, unless v has the minimal age in BC. If this is the case, then flip it with 0.3 probability. Otherwise flip the variable with the second highest net gain \citep{mcallester97}.
\end{itemize}
\item{\bf Ruin-recreate heuristics}
\begin{itemize}
    \item[$h_5$:] A proportion of the variables is randomly reinitialised.
\end{itemize}

\item{\bf Local search heuristics}
\begin{itemize}
    \item[$h_6$:] This is a first-improvement local search. In each iteration, flip a variable selected completely at random.
    \item[$h_7$:] This is a first-improvement local search. In each iteration, flip a randomly selected variable from a randomly selected broken clause.
\end{itemize}

\item{\bf Crossover heuristics}
\begin{itemize}
    \item[$h_8$:] Standard one point crossover on the boolean strings of variables.
    \item[$h_9$:] Standard two point crossover on the boolean strings of variables.
\end{itemize}
\end{description}

\subsection{One Dimensional Bin Packing}
\label{binpacking}
\subsubsection{Description}
\noindent
{\bf Problem formulation}: The classical one dimensional bin packing problem consists of a set of pieces, which must be packed into as few bins as possible. Each piece $j$ has a weight $w_j$, and each bin has capacity $c$. The objective is to minimise the number of bins used, where each piece is assigned to one bin only, and the weight of the pieces in each bin does not exceed $c$. To avoid large plateaus in the search space around the best solutions, we employ an alternative fitness function to the number of bins. A mathematical formulation of the bin packing problem is shown in equation \ref{bp}, taken from \citep{martello90book}.
\begin{align}
\label{bp}
\mbox{Minimise}  \hskip 10mm 	& 	\sum_{i=1}^n y_i \notag \\
\mbox{Subject to} \hskip 10mm & 	\sum_{j=1}^n w_j x_{ij} \le cy_i, 	& i &\in N = \{1,\ldots,n\}, 	 \notag \\
					& 	\sum_{i=1}^n x_{ij} = 1, 		& j &\in N, 			\notag \\
					& 	y_i \in \{0,1\}, 				& i &\in N, 			\notag \\
					& 	x_{ij} \in \{0,1\}, 			& i &\in N, j \in N,
\end{align}
Where $y_i$ is a binary variable indicating whether bin $i$ contains pieces, $x_{ij}$ indicates whether piece $j$ is packed into bin $i$, and $n$ is the number of available bins (and also the number of pieces as we know we can pack $n$ pieces into $n$ bins).

\noindent
{\bf Solution initialisation}: Solutions are initialised by first randomising the order of the pieces, and then applying the `first-fit' heuristic \citep{johnson74}. This is a constructive heuristic, which packs the pieces one at a time, each into the first bin into which they will fit.

\noindent
{\bf Objective function}:  A solution is given a fitness calculated from equation \ref{eq1}, where: $n$ = number of bins, $fullness_i$ = sum of all the pieces in bin \emph{i}, and $C$ = bin capacity. The function puts a premium on bins that are filled completely, or nearly so. It returns a value between zero and one, where lower is better, and a set of completely full bins would return a value of zero.
\begin{equation}
\label{eq1}
Fitness = 1 - \Bigg(\frac{\sum_{i=1}^n (fullness_i / C)^2}{n}\Bigg)
\end{equation}

\noindent
{\bf Instance data}: The ten training instances and their sources are summarised in Table \ref{tab:bpinst}.

\begin{table}
\centering
\caption{Bin packing instances}
\begin{tabular}{c|c|c|c|c}
& name &  source &  capacity &  no. pieces\\
\hline
1 & {\small falkenauer/u1000-00} &  \cite{esicup} & 150 & 1000\\
2 & {\small falkenauer/u1000-01} &  \cite{esicup}& 150 & 1000\\
3 & {\small schoenfieldhard/BPP14}  &  \cite{esicup} & 1000 & 160\\
4 & {\small schoenfieldhard/BPP832} &  \cite{esicup} & 1000 & 160\\
5 & {\small 10-30/instance1} &  \cite{hydeinstances} & 150 & 2000\\
6 & {\small 10-30/instance2} &  \cite{hydeinstances} & 150 & 2000\\
7 & {\small triples1002/instance1} & \cite{hydeinstances} & 1000 & 1002\\
8 & {\small triples2004/instance1} &  \cite{hydeinstances} & 1000 & 2004\\
9 & {\small test/testdual4/binpack0} &  \cite{esicup} & 100 & 5000\\
10 & {\small test/testdual7/binpack0} & \cite{esicup}& 100 & 5000\\

\end{tabular}
\label{tab:bpinst}
\end{table}

\subsubsection{Search Operators}
This  domain contains a total of  8 search operators, some of which are taken from \citep{bai07bpheuristics}.
\begin{description}

\item{\bf Mutational heuristics}
\begin{itemize}
    \item[$h_1$:] Select two different pieces at random, and swap them if there is space. If one of the pieces does not fit into the new bin then put it into an empty bin.
    \item[$h_2$:] This heuristic selects a bin at random from those with more pieces than the average. It then splits this bin into two bins, each containing half of the pieces from the original bin.
    \item[$h_3$:] Remove all of the pieces from the lowest filled bin, and repack them into the other bins if possible, with the best-fit heuristic.
\end{itemize}
\item{\bf Ruin-recreate heuristics}
\begin{itemize}
    \item[$h_4$:] Remove all the pieces from the $x$ highest filled bins, where $x$ is an integer determined by the `intensity of mutation' parameter. Repack the pieces using the best-fit heuristic.
    \item[$h_5$:] Remove all the pieces from the $x$ lowest filled bins, where $x$ is an integer determined by the `intensity of mutation' parameter. Repack the pieces using the best-fit heuristic.
\end{itemize}
\item{\bf Local search heuristics}

These heuristics implement first-improvement local search operators. In each iteration, a neighbour is generated, and it is accepted immediately if it has superior or equal fitness. If the neighbour is worse, then the change is not accepted.
\begin{itemize}
    \item[$h_6$:] A first-improvement local search. In each iteration, select two different pieces at random, and swap them if there is space, and if it will produce an improvement in fitness.
    \item[$h_7$:] A first-improvement local search. Take the largest piece from the lowest filled bin, and exchange with a smaller piece from a randomly selected bin. If there is no such piece that produces a valid packing after the swap, then exchange the first piece with \emph{two} pieces that have a smaller total size. If there are no such pieces then the heuristic does nothing.
\end{itemize}

\item{\bf Crossover heuristics}
\begin{itemize}
    \item[$h_8$:] Exon shuffling crossover \citep{rohlfshagen07}. The bins from both parents are ordered by wasted space, least first. Then all of the mutually exclusive bins are added to the offspring. In the second phase, the remaining bins from the parents are added to the offspring by removing any duplicate pieces.
\end{itemize}
\end{description}

\subsection{Permutation Flow Shop}
\subsubsection{Description}
\noindent
{\bf Problem formulation}: The permutation flow shop problem consists of finding the order in which $n$ jobs are to be processed in $m$ consecutive machines. The jobs are processed in the order machine 1, machine 2,\,\ldots, machine $m$. Machines can only process one job at a time and jobs can be processed by only one machine at a time. No job can jump over any other job, meaning that the order in which jobs are processed in machine 1 is maintained throughout the system. Moreover, no machine
is allowed to remain idle when a job is ready for processing. All jobs and machines are available at time 0. Each job $i$ requires a processing time on machine $j$ denoted by $p_{ij}$.

Given a permutation $\pi=\pi(1),\ldots,\pi(n)$, where $\pi(q)$ is the index of the job assigned in the $q$-th place, a unique schedule is obtained by calculating the starting and completion time of each job on each machine. The starting time $start_{\pi(q),j}$ of the $q$-th job on machine $j$ is calculated as:
\[start_{\pi(q),j} = \max \lbrace
start_{\pi(q),j-1},start_{\pi(q-1),j}\rbrace,\] with
\[start_{\pi(0),j}=0 \hspace{10pt}\mbox{and}\hspace{10pt}
start_{\pi(q),0}= 0,\] and its completion time is calculated as:
\[C_{\pi(q),j}=start_{\pi(q)}+p_{\pi(q),j}.\]

Given a schedule, let $C_i$ be the time when job $i$ finishes its processing on machine $m$. The objective is to find the processing order of $n$ jobs in such a way that the resultant schedule minimises the completion time of the last job to exit the shop, i.e. minimises $\max_iC_i$.

\noindent
{\bf Solution initialisation}: Solutions are created with a randomised version of the widely used NEH algorithm \citep{Nawaz83}, which works as follows. First a random permutation of the jobs is generated. Second, a schedule is constructed from scratch by assigning the first job in the permutation to an empty schedule; the second job is then assigned to places 1 and 2 and fixed where the partial schedule has the smallest makespan; the third job is assigned to places 1, 2 and 3 and fixed to the place where the partial schedule has the smallest makespan, and so on.

\noindent
{\bf Objective function}: The fitness function returns $\max_iC_i$. Representing the completion time of the last job in the schedule.

\noindent
{\bf Instance data}: The ten training instances and their sources are summarised in Table \ref{tab:fsinst}.
\begin{table}
\centering
\caption{Permutation flowshop instances}
\begin{tabular}{c|c|c|c|c}
instance & name & source & no. jobs & no. machines\\
\hline
1 & 100x20/1 &  \cite{Taillard}  & 100 & 20 \\
2 & 100x20/2 & \cite{Taillard}& 100 &  20  \\
3 & 100x20/3 & \cite{Taillard}& 100 &   20 \\
4 & 100x20/4 & \cite{Taillard}& 100 & 20 \\
5 & 100x20/5 & \cite{Taillard}& 100 & 20 \\
6 & 200x10/2 & \cite{Taillard}& 200 & 10 \\
7 & 200x10/3 & \cite{Taillard}& 200&10 \\
8 & 500x20/1 & \cite{Taillard}& 500 & 20\\
9 & 500x20/2 & \cite{Taillard}& 500& 20\\
10 & 500x20/4 & \cite{Taillard}& 500&  20 \\
\end{tabular}
\label{tab:fsinst}
\end{table}

\subsubsection{Search Operators}
A total of 15 search operators are implemented for this problem domain.

\begin{description}
\item {\bf Mutational heuristics}
\begin{itemize}
    \item[$h_1$:] Reinserts a randomly selected job into a randomly selected position in the permutation, shifting the rest of the jobs as required.

    \item[$h_2$:] Swaps two randomly selected jobs in the permutation.

    \item[$h_3$:] Randomly shuffles  the entire permutation.

    \item[$h_4$:] Creates a new solution using NEH (described above) and using  the current permutation to rank the jobs.

    \item[$h_5$:] Shuffles $k$ randomly selected elements in the permutation,
                   where $k=2+\lfloor \alpha\cdot(n-2)\rfloor$,
                   and $\alpha$ is the mutation intensity
                   parameter.
\end{itemize}
\item{\bf Ruin-recreate heuristics}
\begin{itemize}
\item[$h_6$:] Remove $l$, $l = \lfloor \alpha\cdot(n -
            1)\rfloor$, randomly selected jobs and reinsert them
      in an NEH fashion. This heuristic resembles the main
      component of the iterated greedy heuristic proposed by
      \citet{Ruiz07} for the permutation flow shop and later
       by \citet{Ruiz08} for the permutation flow shop with sequence dependent
      setup times.

\item[$h_7$:] Remove $l$, where $l$ is as above, randomly selected jobs, reinsert them in
      an NEH fashion but this time, at every iteration of the NEH
            procedure the best $q$, $q = \lfloor
            \beta\cdot(l-1)\rfloor + 1$, sequences generated so
            far are considered for the reinsertion.
\end{itemize}

\item{\bf Local search heuristics}
\begin{itemize}
    \item[$h_8$:] This is a steepest descent local search. At every     iteration each job is removed from its current position and assigned into all remaining positions. The job is fixed to     the position that leads to the best schedule. This is repeated until no improvement is observed.

    \item[$h_9$:] This is a first improvement local search. At every
    iteration each job is removed from its current position and
    assigned into the remaining positions. This time, if an
    improvement movement is found, this is immediately accepted,
    and the search continues with the next job. This is repeated
    until no improvement is observed.

    \item[$h_{10}$:] This is a random single local search pass.  In this, $r=\lfloor\beta(n-1)\rfloor+1$
    randomly selected jobs are tested (one at a time) on all
    positions and fixed to the best possible place. This is only
    done once.

    \item[$h_{11}$:] This is a first improvement random single local search pass. This is as
    $h_9$ but jobs are assigned to the first place that improves
    the current schedule, i.e. jobs are not necessarily tested in
    all positions. This is only done once.
\end{itemize}

\item{\bf Crossover heuristics}
The following crossover heuristics take two permutations as an input and return a single new permutation as offspring. These operators have been designed for permutation representation problems, including scheduling problems.
\begin{itemize}

\item[$h_{13}$:] Partially mapped crossover (PMX): first proposed by \cite{Goldberg1985}, as a recombination operator for the traveling salesman probem (TSP). It builds an offspring by choosing a subsequence of a tour from one parent and preserving the order and position of as many elements (cities in the case of TSP)  as possible. A subsequence of a tour is selected by randomly choosing two cut points, which serves as boundaries for the swapping operations.

\item[$h_{12}$:] Order crossover (OX): proposed by \cite{Davis1985} for order-based permutation problems. It builds an offspring permutation by choosing a subsequence of a solution from one parent and preserving the relative order of elements from the other parent. The OX operator exploits the property that the relative order of the elements (as opposed to their specific positions) is important.

\item[$h_{14}$:] Precedence preservative crossover (PPX):  independently developed for the vehicle routing problems by \cite{Blanton1993}, and for scheduling problems by \cite{Bierwirth1996}. PPX transmits precedence relations of operations given in two parental permutations to one offspring at the same rate, while no new precedence relations are introduced.

\item[$h_{15}$:] This operators selects a single crossover point and  produces a new permutation by copying all of the elements from one parent, up to the crossover point. Then the remaining elements are copied from the other parent, in the order that they appear.
\end{itemize}
\end{description}

\subsection{Personnel Scheduling }
\label{personnel}
\subsubsection{Description}
\noindent
{\bf Problem formulation}: Most of the personnel scheduling instances could justifiably be labelled as a
new and different problem rather than just a different instance. This is because most instances
contain unique constraints and objectives, not just different instance parameters (such as the number of
employees, shift types, planning period length, constraint priorities etc). The reason
for this variety is that each instance is taken from a different organisation or workplace and each
workplace has its own set of rules and requirements. However, there is clearly a similar structure
between instances and there are some constraints that are nearly always present. For example, cover constraints, holiday requests, maximum and minimum workloads etc. The result of this variety though is that it is arguably impossible to provide a standard mathematical model for `The Personnel Scheduling Problem' and we will not attempt to do so here. However, for more information on the constraints and objectives present in the instances used here (and an integer programming formulation of one of them) we refer the reader to \citet{Curtois2010TR}.

\noindent
{\bf Solution initialisation}: The solution is initialised using local search heuristic $h_5$ which adds shifts to each employee's schedule in a greedy, first improvement manner.

\noindent
{\bf Instance data}: The instances used are listed in Table \ref{tab:psinst}.
\begin{table}
\centering
\caption{Personnel scheduling instances}
\begin{tabular}{c|c|c|c|c|c}

 &  &  &  & shift  & \\
 & name & source & staff &  types & days\\
\hline
1&{\small BCV-3.46.1} & \cite{NRP} & 46 & 3 & 26 \\
2&{\small BCV-A.12.2} & \cite{NRP} & 12 & 5 & 31 \\
3&{\small ORTEC02} & \cite{NRP} & 16 & 4 & 31 \\
4&{\small Ikegami-3Shift-DATA1} & \cite{Ikegami2003} & 25 & 3 & 30 \\
5&{\small Ikegami-3Shift-DATA1.1} & \cite{Ikegami2003} & 25 & 3 & 30 \\
6&{\small Ikegami-3Shift-DATA1.2} & \cite{Ikegami2003} & 25 & 3 & 30\\
7&{\small CHILD-A2} & \cite{NRP} & 41 & 5 & 42 \\
8&{\small ERRVH-A} & \cite{NRP} & 51 & 8 & 42 \\
9&{\small ERRVH-B} & \cite{NRP} & 51 & 8 & 42 \\
10&{\small MER-A} &  \cite{NRP} & 54 & 12 & 42\\
\end{tabular}
\label{tab:psinst}
\end{table}

\subsubsection{Search Operators}

A total of 12 search operators are implemented for this problem domain.

\begin{description}

\item{\bf Mutational heuristics}
\begin{itemize}
\item[$h_1$:] This heuristic randomly un-assigns a number of shifts. The number of shifts un-assigned is proportional to the intensity of mutation parameter.
\end{itemize}

\item{\bf Ruin-recreate heuristics}
The ruin and recreate heuristics implemented are based on the one presented by \citet{NurseBurke08}. The heuristic works by un-assigning all the shifts in one or more randomly selected employees' schedules before heuristically rebuilding them. They are rebuilt by firstly satisfying objectives related to requests to work certain days or shifts and then by satisfying objectives related to weekends. For example min/max weekends on/off, min/max consecutive working or non-working weekends, both days of the weekend on or off etc. Other shifts are then added to the employee's schedule in a greedy fashion (first improvement) attempting to satisfy the rest of the objectives.
\begin{itemize}
\item[$h_2$:] \citet{NurseBurke08} observed that it was best to un-assign and rebuild only 2-6 work patterns at a time (for instances of all sizes). For this reason the first ruin and recreate heuristic un-assigns $x$ schedules where $x$ is calculated using the intensity of mutation parameter as follows:

\begin{center}
\texttt{$x$ = Round(intensityOfMutation * 4) + 2}
\end{center}

\item[$h_3$:] This heuristic provides a larger change to the solution by setting $x$ using:

\begin{center}
\texttt{$x$ = Round(intensityOfMutation * Number of employees in roster)}
\end{center}

\item[$h_4$:] This heuristic creates a small perturbation in the solution by using $x=1$.
\end{itemize}

\item{\bf Local Search Heuristics}
\begin{itemize}
\item[$h_5$:] This is a first improvement local search which adds shifts to employees' schedules.

\item[$h_6$:] This is a first improvement local search which swaps shifts between two different employees. An example of the type of swap this local search may make is shown in Figure \ref{RosterSwap1}. The figure shows a section of a roster showing the
the first ten days of the schedules for four employees: `A', `B', `C' and `D'. The coloured squares labelled `D', `E' and `N' denote three different shifts types (Early, Day and Night)

\item[$h_7$:] This is a first improvement local search which swaps shifts in a single employee's schedule. An example of the type of swap this local search may make is shown in Figure \ref{RosterSwap2}.

\item[$h_8$:] This is based on the ejection chain method described by \citet{NurseBurke2007TR}. The maximum search time for it is set as: the depth of search parameter multiplied by 5 seconds.

\item[$h_9$:] This is another version of the ejection chain method which incorporates a greedy heuristic method for generating entire schedules for single employees. The maximum search time for it is set as: the depth of search parameter multiplied by 5 seconds.

\end{itemize}

\begin{figure}[th]
  \centering
  \includegraphics[scale=0.4]{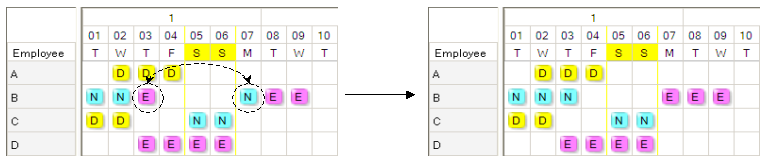}
  \caption{An example of the types of swap made by $h_6$}
  \label{RosterSwap1}
\end{figure}

\begin{figure}[th]
  \centering
  \includegraphics[scale=0.4]{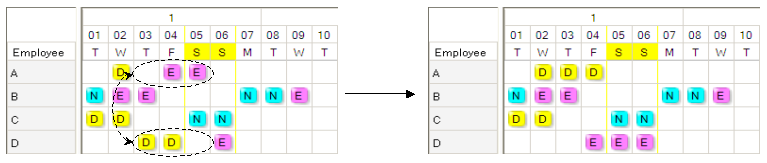}
  \caption{An example of the types of swap made by $h_7$}
  \label{RosterSwap2}
\end{figure}

\item{\bf Crossover heuristics}
\begin{itemize}
\item[$h_{10}$:] This heuristic was presented by \citet{NurseBurke2001b}. It operates by identifying the best $x$ assignments in each parent and making these assignments in the offspring. The best assignments are identified by measuring the change in objective function when each shift is temporarily unassigned in the roster. The best assignments are those that cause the largest increase in the objective function value when they are unassigned. The parameter x ranges from 4-20 and is calculated using the intensity of mutation parameter as below:

\begin{center}
\texttt{$x$ = 4 + round((1 - intensityOfMutation) * 16)}
\end{center}

\item[$h_{11}$:] This heuristic was published in \citep{NurseBurke10}. It creates a new roster by using all the assignments made in the parents. It makes those that are common to both parents first and then alternately selects an assignment from each parent and makes it in the offspring unless the cover objective is already satisfied.

\item[$h_{12}$:] This heuristic creates the new roster by making assignments which are only common to both parents.
\end{itemize}
\end{description}

\section{Algorithms}
\label{algorithms}
This section presents three example algorithms created within the HyFlex software framework. We present these algorithms in order to show the range of algorithms that can be easily implemented in HyFlex. The results of these three algorithms are presented in section \ref{experiments}, to show the diversity of their results across the different problem instances and problem domains. Recall from section \ref{sec:probdomain} that HyFlex problem domains are always implemented as minimisation problems, so a lower fitness is superior.

\subsection{Iterated Local Search}
Iterated local search is a relatively straightforward  algorithm. As often happens with many simple but sometimes very effective ideas, the same principle has been rediscovered multiple times, leading to different names \citep{Baxter1981,Martin1992}. The term {\em iterated local search} was proposed by\cite{Stutzle2002}. The implementation reported here,  first proposed in  in \citep{cec10}, contains a perturbation stage during which a neighborhood move is selected uniformly at random (from the available pool)  and applied to the incumbent solution. This perturbation phase is then followed by an improvement phase, in which all local search heuristics are tested and the one producing the best improvement is used. If the resulting new solution is better than the original solution then it replaces the original solution, otherwise the new solution is simply discarded. This last stage corresponds to a greedy (only improvements) acceptance criterion. The pseudo-code of this iterated local search algorithm  is  shown below (Algorithm  \ref{alg:ils}).

\begin{algorithm}
\caption{{\em Iterated Local Search}. } \label{alg:ils}
\begin{algorithmic}
\STATE $s_0$ = \textrm{GenerateInitialSolution }
\STATE $s^*$ = LocalSearch($s_0$)
\REPEAT
\STATE $s'$ = Perturbation ($s^*$)\vspace{-0.1cm}
\STATE $s^{*'}$ =  LocalSearch($s'$)
        \IF{$f(s^{*'}) < f(s*)$}
            \STATE $s^*$ = $s^{*'}$
    \ENDIF
\UNTIL{time limit is reached}
\vspace{-0.05cm}
\end{algorithmic}
\end{algorithm}

\subsection{Tabu Search Hyper-heuristic with Adaptive Acceptance (TS-AA)}
The functionality of this hyper-heuristic can be split into two parts, the heuristic selection mechanism and the move acceptance criteria. The pseudocode for TS-AA can be seen in Algorithm \ref{alg:tsaa}.

\noindent
{\bf Heuristic selection mechanism for TS-AA}:
This hyper-heuristic implements the heuristic selection mechanism proposed in \citep{burke03tabuhh}.The algorithm maintains a value for each of the problem-specific heuristics, excluding the crossover type heuristics. The crossover heuristics are not used at all by this hyper-heuristic. The heuristic's value represents how well it has performed recently, and all heuristics have a value of zero at the beginning of the search. The mechanism also incorporates a dynamic tabu list of problem-specific heuristics that are temporarily excluded from the available heuristics.

At each iteration, the heuristic with the highest value is selected (breaking ties randomly), from those not in the tabu list. Therefore, the heuristics which have performed well recently will be chosen more often. If the heuristic finds a better solution, then its value is increased. If it finds a worse solution, its value is decreased.

\noindent
{\bf Acceptance criterion for TS-AA}:
The acceptance criterion accepts all improving solutions. Other solutions are accepted with a probability $\beta$. which changes depending on whether the search appears to be progressing or stuck in a local optimum. The $\beta$ value begins at zero, thus, initially, it is  an accept-only-improving strategy. However, if the solution does not improve for 0.1 seconds, then $\beta$ is increased by $5\%$, making it more likely that a worse solution is accepted. It is increased to $10\%$ if there is no further improvement in the next 0.1 seconds. Conversely, if the search is progressing well, with no decrease in fitness in the last 0.1 seconds, then $\beta$ is reduced by $5\%$, making it less likely for a worse solution to be accepted. These modifications are intended to help the search navigate out of local optima, and to focus the search if it is progressing well.

\begin{algorithm}
\caption{The Tabu Search Hyper-heuristic with Adaptive Acceptance}
\label{alg:tsaa}
\begin{algorithmic}
\STATE Create a initial solution $s$
\STATE Initialise the value of each heuristic to 0
\STATE $\alpha = 1$
\STATE $\beta = 0$
\STATE $t=$ tabu tenure = number of heuristics-1
\REPEAT
\STATE Create a copy of the current solution: $s' \leftarrow s$
\STATE $H =$ heuristic with highest value
\STATE apply $H$ to  $s'$
\IF[the new solution is superior]{$func(s') < func(s)$}
    \STATE $increaseValue(H, \alpha)$
\ELSIF[the new solution is worse]{$func(s') < func(s)$}
    \STATE empty the tabu list
    \STATE $decreaseValue(H, \alpha)$
    \STATE add $H$ to the tabu list
\ELSIF {$func(s') = func(s)$}
    \STATE add $H$ to the tabu list
    \STATE release heuristics in tabu list for longer than $t$ iterations
\ENDIF

\IF{$func(s') < func(s)$}
	\STATE $s \leftarrow s'$
\ELSE
	\IF{$random[1,100] < \beta$}
        \STATE $s \leftarrow s'$	
    \ENDIF
\ENDIF

\IF{0.1s since last improvement}
    \STATE \COMMENT{make it more likely to accept worse solutions}
    \STATE $\beta \leftarrow \beta + 5$
\ENDIF
\IF{0.1s since last decrease in fitness}
    \STATE \COMMENT{make it less likely to accept worse solutions}
    \STATE $\beta \leftarrow \beta - 5$
\ENDIF

\UNTIL{time limit is reached}
\end{algorithmic}
\end{algorithm}

\subsection{Memetic Algorithm}

This algorithm illustrates a population based approach implemented with HyFlex. It represents a steady-state evolutionary algorithm that incorporates multiple memes (a memetic algorithm). The pseudocode is given in Algorithm \ref{alg:mem}. First a population of 10 solutions is generated, each one initialised with the \texttt{initialiseSolution()} method provided by each problem domain. Two solutions are selected with a binary tournament method, and then a crossover type heuristic (selected uniformly at random from the available set) is applied to produce one offspring.

With 0.1 probability, the offspring is perturbed with a mutation heuristic (selected uniformly at random from the available set). Then the solution is further modified with either a local search heuristic or a ruin-recreate heuristic, chosen with a 0.5 probability (also selected each uniformly at random). If the new solution is equal to or better than the worst of the parents, then the offspring replaces it.

\begin{algorithm}
\caption{Memetic algorithm} \label{alg:mem}
\begin{algorithmic}
\STATE $population$ = Create a initial population of 10 solutions
\REPEAT
\STATE s1 = binaryTournament(population);
\STATE s2 = binaryTournament(population);
\STATE $h$ =   randomly selected crossover heuristic
\STATE $s'$ = applyheuristic($h$, s1, s2);
\STATE Apply randomly selected mutation heuristic to  $s'$
\IF {$rand < 0.5$}
    \STATE $h$ = randomly selected local search heuristic
\ELSE
    \STATE $h$ = randomly selected ruin-recreate heuristic
\ENDIF
\STATE applyheuristic($h$, $s'$);
    \IF{$f(s1)$ worse than $f(s2)$}
        \STATE $s1 \leftarrow s'$
    \ELSE
        \STATE $s2 \leftarrow s'$
    \ENDIF
\UNTIL{time limit is reached}
\end{algorithmic}
\end{algorithm}

\section{Experiments and Results}
\label{experiments}
This section compares the three algorithms described in section \ref{algorithms} implemented with HyFlex. Exactly the same algorithms are used for each domain and instance. No domain-specific (or instance-specific) tuning process is applied. The goal is not to determine which is the best performing algorithm, but instead to illustrate the behaviour of different algorithmic designs in HyFlex.  The 10 training instances for each domain, as described in section \ref{hyflex} (Tables \ref{tab:satinst}-\ref{tab:psinst}), were considered. For each instance and algorithm, 5 runs were conducted, each lasting 10 CPU minutes. This experimental setup resembles that designed for the CHeSC competition. The experiments were conducted on a PC (running Windows XP) with a 2.33GHz Intel(R) Core(TM)2 Duo CPU and 2GB of RAM. The following subsections present our results from three different perspectives: ordinal data analysis (\ref{sec:one}), distribution of best objective function values on one selected instance per domain (\ref{sec:two}), performance behaviour over time on one example instance of the bin packing domain (\ref{sec:three}).

\subsection{Borda count}\label{sec:one}

Ordinal data analysis methods can be applied to compare alternative search algorithms or metaheuristics  \citep{Talbi2009}. This approach is adequate because our empirical study considers different domains and instances with varied magnitudes and ranges of the objective values. Let us assume that $m$ instances (considering all the domains) and $n$ competing algorithms in total are considered. For each experiment (instance) an ordinal value $o_k$ is given representing the rank of the algorithm compared to the others ($1 \leq o_k \leq n$). Ordinal  methods aggregate and summarise $m$ linear orders $o_k$ into a single linear order $O$. We use here a straight forward ordinal aggregation method know as the {\em Borda count} voting method (after the French mathematician Jean-Charles de Borda, who first proposed it in 1770). An algorithm having a rank $o_k$ in a given instance is simply given $o_k$ points, and  the total score of an algorithm is the sum of its ranks $o_k$ across the $m$ instances. The methods are, therefore, compared according to their total score, with the smallest score representing the best performing algorithm. In our comparative study, the number of instances, $m$, is 40 (10 for each domain). Therefore, for a given domain the best possible score is 10, while the best possible total score (considering all the domains) is 40. The ranks were calculated using as a metric the median of the best objective functions obtained across the 5 runs per instance.

Table \ref{tab:borda} shows the total Borda scores for the three competing algorithms, including the total scores per domain. Notice that although {\em TS-AA} produces the best scores in two domains: MAX-SAT and permutation flow shop; the {\em ILS} algorithm obtains the best overall scores, although by a minimal difference. Tables \ref{tab:satborda}-\ref{tab:psborda} show the Borda count (ranks) for each instance on the four domains, where 1 represents the best rank. These tables are useful to assess how homogeneous the results are for the ten instances on each domain. For example, for permutation flow shop and personnel scheduling (Tables \ref{tab:fsborda}-\ref{tab:psborda}) a single algorithm is consistently ranking 3rd, whereas this is not the case for MAX-SAT and bin packing (Tables \ref{tab:satborda}-\ref{tab:bpborda}).

\begin{table}
\centering
\caption{Borda count results for all domains}
\begin{tabular}{c|c|c|c}
Domain &  {\em TS-AA} & {\em ILS} & $MA$\\
\hline
MAX-SAT & {\bf 12} & 27 & 21\\
1D Bin Packing & 24 & {\bf 17} & 19\\
Permutation Flow Shop & 30 & 17 & {\bf 13}\\
Personnel Scheduling & {\bf 13} & 16 & 30\\
\hline
Total & 79 & {\bf 77} & 83\\
\end{tabular}
\label{tab:borda}
\end{table}

\begin{table}
\centering
\begin{minipage}{0.45\textwidth}
\centering
\caption{Borda count results for MAX-SAT}
\label{tab:satborda}
\begin{tabular}{c|c|c|c}
MAX-SAT & {\em TS-AA} & {\em ILS} & $MA$\\
\hline
Instance1 & 2  & 3  & 1\\
Instance2 & 2  & 3 & 1\\
Instance3 & 1  & 3 & 2\\
Instance4 & 1  & 2 & 3\\
Instance5 & 1 &  2 & 3\\
Instance6&  1  & 2  & 3\\
Instance7 & 1  & 3 & 2\\
Instance8 & 1  & 3 & 2\\
Instance9 & 1  & 3 & 2\\
Instance10 & 1  & 3 & 2\\
\hline
Total & {\bf 12} & 27 & 21\\
\end{tabular}
\end{minipage}
\begin{minipage}{0.05\textwidth}
\verb+ +\\
\end{minipage}
\begin{minipage}{0.45\textwidth}
\centering
\caption{Borda count results for 1D Bin Packing}
\label{tab:bpborda}
\begin{tabular}{c|c|c|c}
Bin Packing & {\em TS-AA} & {\em ILS} & $MA$\\
\hline
Instance1 & 3  & 1  & 2\\
Instance2 & 3  & 1 & 2\\
Instance3 & 3  & 2 & 1\\
Instance4 & 2  & 3 & 1\\
Instance5 & 2 &  1 & 3\\
Instance6&  3  & 1  & 2\\
Instance7 & 3  & 1 & 2\\
Instance8 & 3 & 1 & 2\\
Instance9 & 1  & 3 & 2\\
Instance10 & 1  & 3 & 2\\
\hline
Total &24 & {\bf 17} & 19\\
\end{tabular}
\end{minipage}
\end{table}

\begin{table}
\centering
\begin{minipage}{0.45\textwidth}
\centering
\caption{Borda count results for permutation flowshop}
\label{tab:fsborda}
\begin{tabular}{c|c|c|c}
Flow Shop & {\em TS-AA} & {\em ILS} & $MA$\\
\hline
Instance1 & 3	& 2	& 1\\
Instance2 & 3	& 1	& 2\\
Instance3 & 3	& 2	& 1\\
Instance4 & 3	& 2	& 1\\
Instance5 & 3	& 2	& 1\\
Instance6 & 3	& 2	& 1\\
Instance7 & 3	& 2	& 1\\
Instance8 & 3	& 1	& 2\\
Instance9 & 3	& 1	& 2\\
Instance10 &3	& 2	& 1\\
\hline
Total &  30 & 17 & {\bf 13}\\
\end{tabular}
\end{minipage}
\begin{minipage}{0.05\textwidth}
\verb+ +\\
\end{minipage}
\begin{minipage}{0.45\textwidth}
\centering
\caption{Borda count results for personnel scheduling}
\label{tab:psborda}
\begin{tabular}{c|c|c|c}
Personnel Sched. & {\em TS-AA} & {\em ILS} & $MA$\\
\hline
Instance1 & 1 & 2 & 3\\
Instance2 & 2	& 1& 3\\
Instance3 & 1	& 1 & 3\\
Instance4 & 2	& 1	& 3\\
Instance5 & 1	& 2	& 3\\
Instance6 & 2	& 1	& 3\\
Instance7 & 1	& 2	& 3\\
Instance8 & 1	& 2	& 3\\
Instance9 & 1	& 2	& 3\\
Instance10 & 1 &	2	& 3\\
\hline
Total &{\bf 13} &  16 & 30\\
\end{tabular}
\end{minipage}
\end{table}

\subsection{Distribution of the best objective function values}\label{sec:two}
\begin{figure} [!ht]
\centering
\begin{minipage}{0.45\textwidth}
\centering
\caption{Distribution of objective function values for the MAX-SAT instance 1: {\small contest02-Mat26.sat05-457.reshuffled-07}}
\label{fig:satplot}
\includegraphics[width=1.0\textwidth]{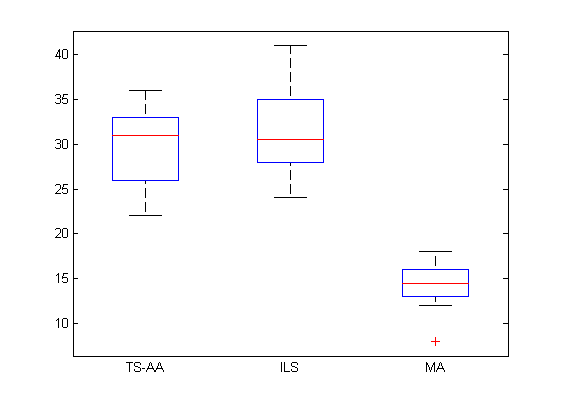}
\end{minipage}
\begin{minipage}{0.05\textwidth}
\verb+ +\\
\end{minipage}
\begin{minipage}{0.45\textwidth}
\centering
\caption{Distribution of objective function values for the bin packing instance 1: {\small falkenauer/u1000-00}}
\label{fig:bpplot}
\includegraphics[width=1.0\textwidth]{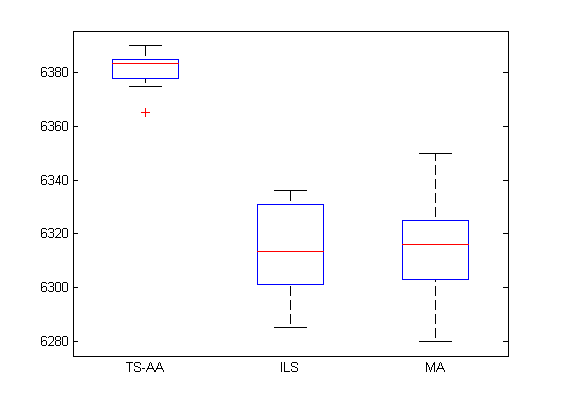}
\end{minipage}
\end{figure}

\begin{figure} [!ht]
\centering
\begin{minipage}{0.45\textwidth}
\centering
\caption{Distribution of objective function values for  the permutation flow shop instance 1: {\small 100x20/1}}
\label{fig:fsplot}
\includegraphics[width=1.0\textwidth]{fsplot.png}
\end{minipage}
\begin{minipage}{0.05\textwidth}
\verb+ +\\
\end{minipage}
\begin{minipage}{0.45\textwidth}
\centering
\caption{Distribution of objective function values for the personnel scheduling instance 1: {\small BCV-3.46.1}}
\label{fig:psplot}
\includegraphics[width=1.0\textwidth]{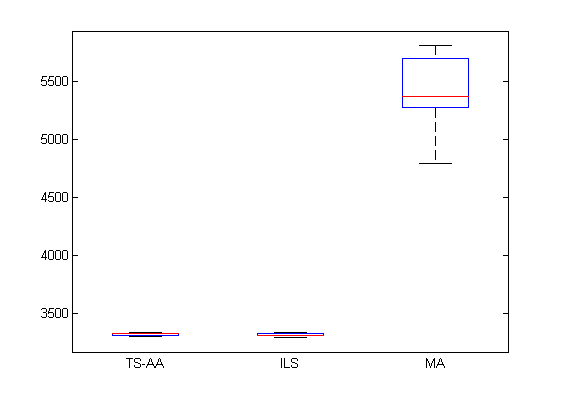}
\end{minipage}
\end{figure}

In addition to the Borda aggregation method presented above, the boxplots shown in figures \ref{fig:satplot}-\ref{fig:psplot} illustrate the magnitude and distribution of the best objective values (at the end of the run) for a selected instance of each domain. Each figure represents the result of 10 runs from each algorithm. Arbitrarily we selected instance number 1 from each domain, but similar distributions of results can be observed in the other instances.

From figures \ref{fig:satplot}-\ref{fig:psplot}, it can be observed that the performance of the three algorithms differs significantly over the four problem instances. For example, in the max-sat instance (figure \ref{fig:satplot}) the memetic algorithm (MA) performs the best, while it performs the worst in personnel scheduling instance(figure \ref{fig:psplot}). The tabu search hyper-heuristic (TS-AA) clearly performs the worst on the instances of bin packing and flow shop (Figures \ref{fig:bpplot}-\ref{fig:fsplot}), but performs the best on the personnel scheduling instance. The scale of Figure \ref{fig:psplot} means that it is difficult to see the difference between TS-AA and ILS. This is because the personnel scheduling domain applies penalties to solution that violates the constraints, and the memetic algorithm produced poor solutions in this instance.

In summary, these boxplots show that it is challenging to design an algorithm which operates well over all the problem domains. When an algorithm improves on one domain, its solution quality may reduce on another domain. This can also be true in a single domain, when an algorithm improves on a particular problem instance, and its performance reduces on other instances of that domain. The challenge is to design online learning mechanisms that can adapt on the fly, and thus select the most adequate heuristic at each decision step,  using the feedback gathered from the search process.

\subsection{Progress of algorithms during a run}\label{sec:three}
\begin{figure}
\centering
\begin{minipage}{0.8\textwidth}
\centering
\caption{1D Bin Packing trace on instance 1, showing the progress of the three example algorithms over 10 minutes}
\label{fig:bptrace}
\includegraphics[width=1.0\textwidth]{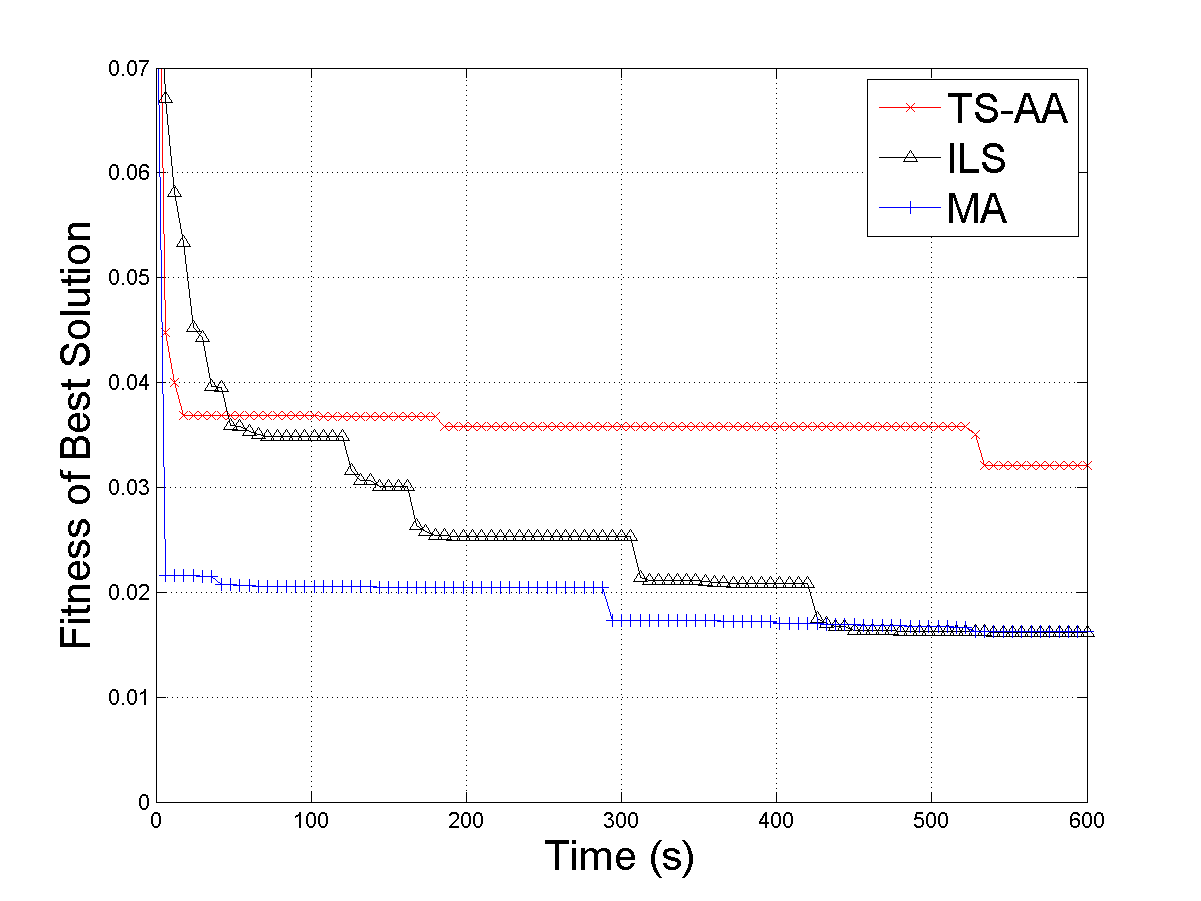}
\end{minipage}
\end{figure}

Figure \ref{fig:bptrace} shows the progress of the three algorithms during one 10 minute run on instance 1 of Bin Packing. A lower fitness value represents a better solution. This information is easily available from within HyFlex by calling the \texttt{getFitnessTrace()} method, and it is automatically recorded during the run. They show that the performance of the algorithms can differ greatly depending on how long they are left to run. Iterated local search (ILS) and the memetic algorithm (MA) both finish the run at approximately the same fitness. However, the memetic algorithm finds better quality solutions more quickly. The tabu search hyper-heuristic (TS-AA) begins the run by finding better solutions than ILS, but TS-AA stagnates, and by the end of the run ILS has found a better solution. This ability to easily obtain useful information for analysis is another way that HyFlex can save a significant amount of time for researchers.

\section{Conclusions}
\label{conclusions}

This paper has presented and described the HyFlex software framework for the development of cross-domain heuristic search methodologies. HyFlex provides multiple problem domains, each containing a set of problem instances and search operators  to apply. Therefore, it represents a novel extension of the notion of  benchmark  for combinatorial optimisation,  with which cross-domain algorithms can be easily developed, and reliably compared. Researchers from different communities and themes within computer science, artificial intelligence and operational research, can potentially benefit from  HyFlex, as it provides a common benchmark in which to test the performance and behavior of single-point and population-based  self-configuring search heuristics. When  using HyFlex, researchers can concentrate their efforts on designing their adaptive methodologies, rather than implementing the required set of problem domains.

This paper describes the architecture of HyFlex, including examples of how to create and run hyper-heuristics within the framework. The four problem domains are presented and discussed, and three example hyper-heuristics are analysed, with their results. The results show that the hyper-heuristics all have differing performances on the four problem domains. No one algorithm is superior to the other two algorithms on all four problem domains. Although these are not state-of-the-art adaptive algorithms, the results suggest  that there is still considerable scope for future research when desiging adaptive and self-configuring algorithms that can learn  from the search process and select the most suitable search operators.

There is currently ample evidence that HyFlex is useful to the research community, due to the number of researchers which are currently employing it for their research and teaching. The HyFlex framework was made publicly available in August 2010. In May 2011, the software had been downloaded over 460 times, and the associated web-pages describing it had been visited over 11,844 times. The community has also responded well to a call for participation in the International Cross-domain Heuristic Search Challenge (CHeSC), which would not be possible without the HyFlex software. In May 2011, the competition had 43 registered participants and teams from 23 different countries.

HyFlex  can be extended to include new domains, additional instances and operators in existing domains, and multi-objective and dynamic problems. The current software interface can also be extended to incorporate additional feedback information from the domains to guide the adaptive search controllers. It is our vision that the HyFlex framework will continue to facilitate and increase international interest in developing adaptive heuristic search methodologies, that can find wider application in practice.

\small


\end{document}